\documentclass[lettersize,journal]{IEEEtran}
\usepackage{amsmath,amsfonts}
\usepackage{array}
\usepackage{float}
\usepackage[caption=false,font=normalsize,labelfont=sf,textfont=sf]{subfig}
\usepackage[ruled,lined]{algorithm2e}
\usepackage{utfsym}
\usepackage{textcomp}
\usepackage{stfloats}
\usepackage{url}
\usepackage{verbatim}
\usepackage{graphicx}
\usepackage{cite}
\usepackage{hyperref}
\begin{document}

\title{Closing the Confusion Loop: CLIP-Guided Alignment for
Source-Free Domain Adaptation}

\author{Shanshan Wang, Ziying Feng, Xiaozheng Shen, Xun Yang*, Pichao Wang, Zhenwei He, Xingyi Zhang~\IEEEmembership{Fellow,~IEEE}
\thanks{Shanshan Wang and Ziying Feng are with the State Key Laboratory of Opto-Electronic Information Acquisition and Protection Technology, Institutes of Physical Science and Information Technology, Anhui University, Hefei 230601, China. (e-mail: wang.shanshan@ahu.edu.cn, q23301278@stu.ahu.edu.cn).}

\thanks{Xiaozheng Shen are with the State Key Laboratory of Opto-Electronic Information Acquisition and Protection Technology, Institutes of Physical Science and Information Technology, Anhui University, Hefei 230601, China. (e-mail:  shenxiaozheng1229@gmail.com).}

\thanks{*Xun Yang~ (Corresponding author) is with the Department of Electronic Engineering and Information Science, School of Information Science and Technology, University of Science and Technology of China, Hefei 230026, China (e-mail: hfutyangxun@gmail.com).}

\thanks{Pichao Wang is with amazon, U.S.A. (e-mail: pichaowang@gmail.com).}

\thanks{Zhenwei He  is with the College of Computer Science and Engineering, Chongqing University of Technology, Chongqing 400044, China (e-mail: hzw@cqut.edu.cn).}

\thanks{Xingyi Zhang is with the Key Laboratory of Intelligent Computing and Signal Processing, Ministry of Education, and the School of Computer Science and Technology, Anhui University, Hefei 230601, China (e-mail: xyzhanghust@gmail.com).}
}

\markboth{IEEE Transactions}%
{Shell \MakeLowercase{\textit{et al.}}: A Sample Article Using IEEEtran.cls for IEEE Journals}


\maketitle

\begin{abstract}
Source-Free Domain Adaptation (SFDA) tackles the problem of adapting a pre-trained source model to an unlabeled target domain without accessing any source data, which is quite suitable for the field of  data security. Although recent advances have shown that pseudo-labeling strategies can be effective, they often fail in fine-grained scenarios due to subtle inter-class similarities. A critical but underexplored issue is the presence of \textit{asymmetric and dynamic class confusion}—where visually similar classes are unequally and inconsistently misclassified by the source model. Existing methods typically ignore such confusion patterns, leading to noisy pseudo-labels and poor target discrimination.
To address this, we propose \textbf{CLIP-Guided Alignment~(CGA)}, a novel framework that explicitly models and mitigates class confusion in SFDA. Generally, our method consists of three parts: (1) MCA: detects first directional confusion pairs by analyzing the predictions of the source model in the target domain; (2) MCC: leverages CLIP to construct confusion-aware textual prompts (\textit{e.g.}, ``a truck that looks like a bus''), enabling more context-sensitive pseudo-labeling; and (3) FAM: builds confusion-guided feature banks for both CLIP and the source model and aligns them using contrastive learning to reduce ambiguity in the representation space.
Extensive experiments on various datasets demonstrate that CGA consistently outperforms state-of-the-art SFDA methods, with especially notable gains in confusion-prone and fine-grained scenarios. Our results highlight the importance of explicitly modeling inter-class confusion for effective source-free adaptation. Our code can be find at \href{https://github.com/soloiro/CGA}{https://github.com/soloiro/CGA}
\end{abstract}

\begin{IEEEkeywords}
Source-free domain adaptation, Vision-language models, Prompt learning, Image classification.
\end{IEEEkeywords}


\section{Introduction}

 \IEEEPARstart{U}nsupervised domain Adaptation (UDA) aims to transfer knowledge from a labeled source domain to an unlabeled target domain, enabling models to generalize across domain shifts without requiring manual annotation in the target domain~\cite{wang2023disentangled}~\cite{ganin2016domain}~\cite{kang2019contrastive}. Conventional UDA methods typically assume full access to source data and jointly optimize source and target representations~\cite{wang2024equity}~\cite{10857978}~\cite{11142945}. 
 
 However, in many real-world scenarios, such as those involving privacy, security, or intellectual property, direct access to source data may be legally or practically infeasible~\cite{mitsuzumi2024understanding}.
\textit{e.g.}, in applications involving medical imaging, personal biometrics, or financial records, sharing raw data from the source domain - even for model adaptation purposes - could violate confidentiality or privacy. 
Such cases have led to the emergence of \textit{Source-Free Domain Adaptation} ~(SFDA)~\cite{li2020model}~\cite{pei2023uncertainty}~\cite{kundu2022concurrent}~\cite{shen2023balancing}, a more realistic and secure setting in which only a pre-trained source model is provided for target adaptation, without access to the original source data. This approach significantly reduces the attack surface for privacy breaches while maintaining the practical benefits of domain adaptation~\cite{tang2023consistency}~\cite{boudiaf2023search}~\cite{zhang2023rethinking}.

\begin{figure}[t]
\begin{minipage}{\linewidth}
    \centering
    \includegraphics[scale=0.65]{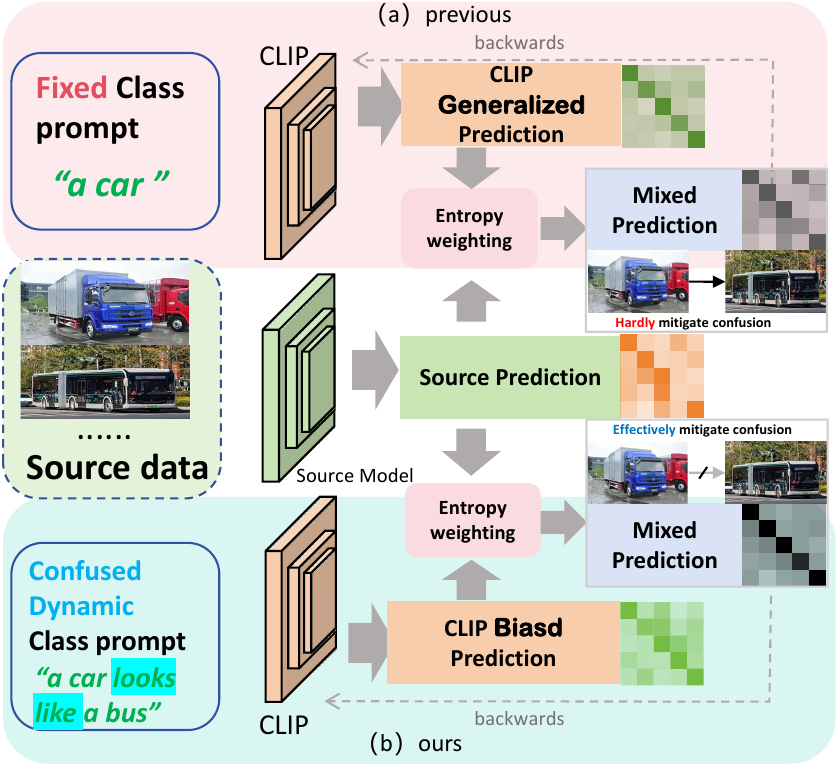}
    \end{minipage}
    
    
    \caption{During conventional mutual learning, cross-domain misconceptions from the source model contaminate the generalized knowledge in CLIP. Our method which employ confusing-class prompts to induce biased knowledge in CLIP can prevent contamination from cross-domain misconceptions.  }
    \label{figure_motivation}

\end{figure}



Despite promising progress, SFDA remains highly challenging due to its ill-posed nature. Without source supervision, the adaptation process is largely based on the predictions of the source model in the target domain. Most existing SFDA methods adopt a pseudo-labeling paradigm, where soft predictions from the source model are refined through clustering or self-training~\cite{grandvalet2004semi}. These approaches typically rely on the cluster assumption, that is, the belief that the target features form well-separated clusters aligned with semantic classes. However, this assumption often fails in fine-grained settings, where categories exhibit only subtle visual differences. In such cases, noisy pseudo-labels arising from inter-class ambiguity can severely degrade model adaptation~\cite{cascante2021curriculum}.

A core yet underexplored reason for this degradation lies in the \textit{asymmetric and dynamic nature of class confusion} in the target domain~\cite{tang2022invariant}. Specifically, certain classes are inherently more prone to confusion due to visual similarity (\textit{e.g.}, {truck} \textit{V.S.} {bus}), and on specific datasets the confusion is often one-directional (\textit{e.g.}, a \textit{truck} might be misclassified as a \textit{bus}, but rarely vice versa due to disparities in training data volume, ). Existing SFDA methods largely overlook this asymmetry and typically treat all inter-class relations equally, thus ignoring the nuanced, instance-specific confusion structure that evolves during adaptation~\cite{zhang2023class}. This mismatch not only leads to incorrect pseudo-labels, but also distorts the learned representation space.
Furthermore, confusion between classes in the target domain could even pose a potential privacy risk~\cite{melis2019exploiting}~\cite{fredrikson2015model}, increasing the urgency to address this issue. 


To alleviate this problem, we propose a principled framework called \textbf{\textit{CLIP-Guided Alignment (CGA)}} to explicitly detect, model and resolve the class confusion of the source model in SFDA setting. Instead of passively tolerating noisy pseudo-labels, CGA places class confusion at the center of the adaptation process and addresses it through three tightly coupled stages: \textit{confusion perception}, \textit{confusion representation}, and \textit{confusion alignment}.

\noindent\textbf{(1) Confusion perception: identification of directional confusion patterns.}  
We begin by analyzing the soft predictions of the source model over the entire target dataset, constructing a directed and asymmetric confusion graph that reveals which pairs of classes are frequently confused. For example, we may find that many ``truck'' samples are misclassified as ``bus'', forming a high-weighted edge in the confusion graph. Unlike prior works that rely on fixed or symmetric relations, our approach dynamically captures these confusion patterns in a data-driven and evolving fashion, adapting to the model's behavior throughout training.

\noindent\textbf{(2) Confusion representation: encoding ambiguity with CLIP-guided prompts.} To improve the quality of supervision under class ambiguity, we leverage the generalization power of Contrastive Language–Image Pre-training~ (CLIP)~\cite{radford2021learning}. For each pair of confusion, we construct hybrid textual prompts such as ``a \textit{truck} that looks like a \textit{bus}'' to represent fine-grained visual ambiguity. These prompts form a set of \textit{confusion-aware textual prototypes} in CLIP’s semantic space. By projecting target images onto these prototypes, we can obtain refined, context-aware pseudo-labels that not only explicitly model class ambiguity—far more reliable than standard CLIP zero-shot predictions or raw source model outputs but also effectively prevent contamination of the CLIP model by pseudo-labels containing cross-domain misconceptions exported from the source model, as shown in Fig.\ref{figure_motivation}.

\noindent\textbf{(3) Confusion alignment: transferring semantics via contrastive feature matching.}  
To resolve confusion at the representation level, we first construct two confusion-aware feature banks, one from the source model and one from CLIP, by aggregating feature centroids over each confusion-aware pseudo-class. We then perform contrastive alignment between these banks to implicitly transfer CLIP's semantic priors to the feature space of the source model. This alignment step reduces inter-class overlap in the target domain, producing more discriminative and semantically structured representations without requiring any source data.

Together, these three stages form a closed loop pipeline: class confusion is first \textit{perceived}, then \textit{encoded} into refined supervision signals, and finally \textit{resolved} via feature-space alignment. This design enables CGA to systematically mitigate the dominant source of adaptation errors in SFDA between visually similar classes.


Our main contributions are summarized as follows.
\begin{itemize}

    \item We uncover the problem of \textbf{ dynamic, asymmetric inter-class confusion} in SFDA, and demonstrate its detrimental effect on pseudo-labeling and feature learning in fine-grained settings.

    \item We propose \textbf{CGA}, a novel framework that addresses class confusion through three synergistic stages: confusion detection, CLIP-guided prompt encoding and alignment of features sensitive to confusion, without requiring access to source data.

    \item We design a \textbf{multi-prototype prompting strategy} that could transform CLIP into a context-sensitive supervision assistant, allowing pseudo-labels to reflect nuanced class ambiguity via textual hybridization (\textit{, e.g.}, ``a truck that looks like a bus'') by adding confusion class text.

    \item We introduce a \textbf{confusion-aware contrastive alignment} mechanism that bridges the feature space of the source model with CLIP’s semantic priors, leading to more structured and discriminative target representations.

    \item CGA establishes new state-of-the-art performance on four widely used SFDA benchmarks, with significant improvements in confusion-prone scenarios, validating the necessity and effectiveness of explicitly modeling class confusion.
\end{itemize}

\section{RELATED WORK}
\subsection{Source-Free Domain Adaptation.}
Recently, the primary direction in SFDA methods has been the use of pseudo-labels for model self-training~\cite{yang2022attracting}~\cite{karim2023c}~\cite{ding2022source}.
They were based on the cluster assumption, which assumed that the source model has some generalizability in the target domain, which means that samples with the same label should be close in the feature space~\cite{yang2023dc}. Therefore, fine-tuning can be conducted by exploring its features or outputs on the target data. For example, CoWA-JMDS~\cite{lee2022confidence} obtained pseudo-labels using Gaussian mixture models (GMM) and further mined the knowledge in the target domain using weighted mixture. PLUE~\cite{litrico2023guiding} estimated the confidence of the samples by calculating the probability entropy of neighboring samples in the feature space.
CRS~\cite{zhang2023class} considered calculating category similarity using the classifier weights of the source model. However, the similarity relationships it derived are bidirectional and fixed and did not account for the bidirectional and dynamically evolving nature of class confusion relationships during training.
Instead, we introduce dynamic confusion estimation into self-training to biased leverage external knowledge from CLIP to improve pseudo-labels.


\subsection{Multimodal Pre-trained Neural Network.}
Multimodal pre-trained models, with CLIP~\cite{radford2021learning} as a prominent example, have been extensively utilized in DA due to their remarkable zero-shot capabilities and great generalization power~\cite{yang2022video}. Typically, these models were customized through textual and visual prompts~\cite{zhou2022learning}~\cite{jia2022visual} or adapter~\cite{gao2024clip}, enabling them to achieve enhanced performance by modifying a few learnable parameters. 
DAPL~\cite{ge2023domain}, fine-tuned CLIP using prompt techniques to effectively leverage domain-specific features, aided in domain generalization. Other UDA methods, such as PDA~\cite{bai2024prompt}, have proposed a prompt-based distribution alignment approach that reduced the distribution gap between the source and target domains using a multimodal strategy, thus improving the effectiveness of UDA.

In the context of SFDA, the application of multimodal pre-trained models for task optimization is still relatively limited. DIFO~\cite{tang2024source} has pioneered the exploration of the potential of the ViL model in SFDA, customized the CLIP model and employed the method to extract specific effective knowledge from multimodal data, thus enhancing the performance of SFDA. ProDe~\cite{tang2024proxy} investigated the noise impact of CLIP on the source model and proposed a proxy denoising approach that effectively leverages CLIP's knowledge. 
Furthermore, although feature alignment was a common and effective approach in DA~\cite{wilson2023calda}~\cite{wang2024dual}~\cite{10841964},
few methods can effectively optimize the feature space of the source model using CLIP, due to the significant gap between the CLIP and the source model in the feature space. Oriented by classification probabilities, 
we construct two confusion-aware class center feature banks and implicitly align these banks to refine the feature space of the source model with CLIP’s knowledge.

\begin{figure*}[t]
\begin{minipage}{\linewidth}
    \centering
    \includegraphics[scale=0.55]{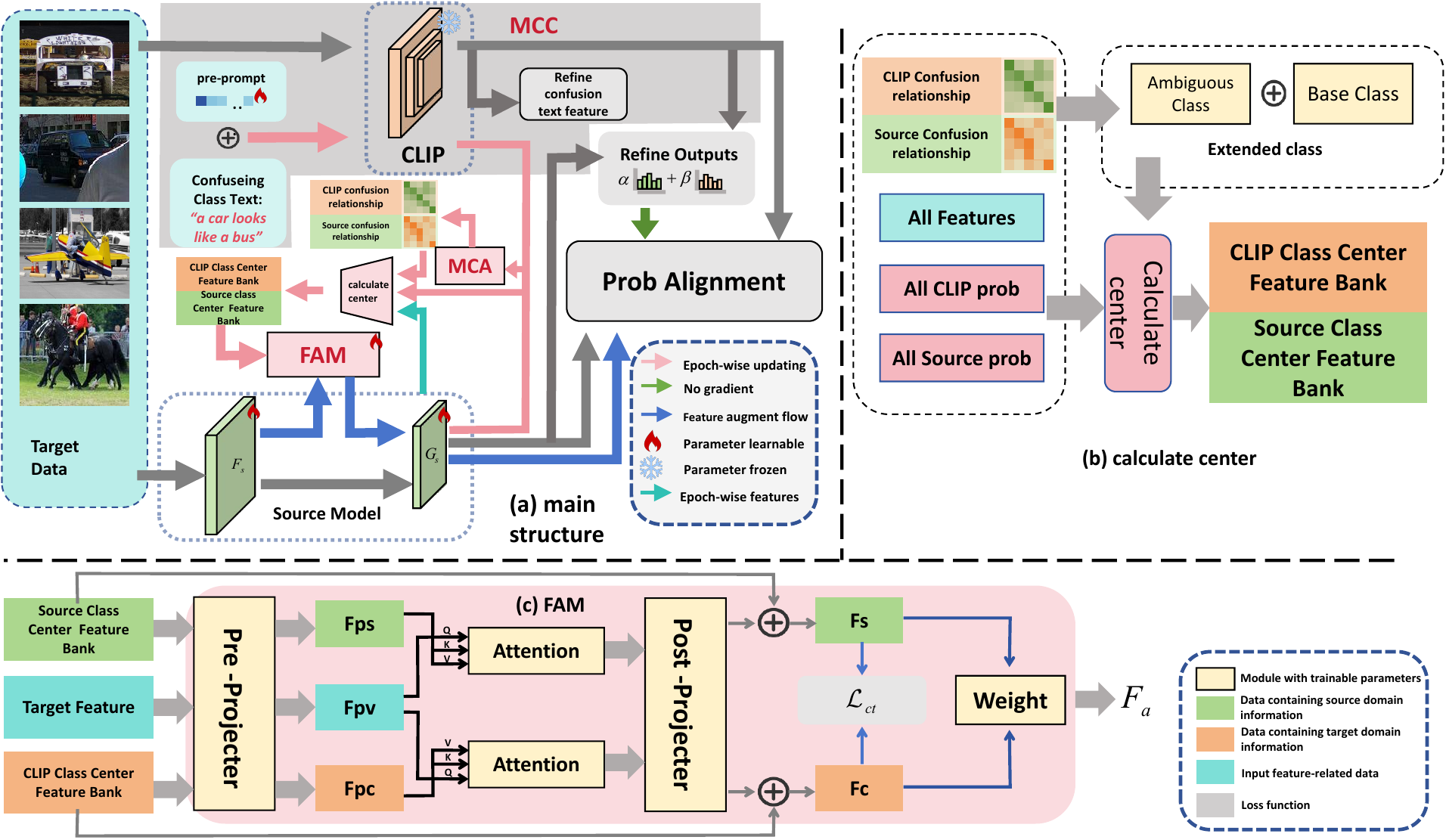}
    \end{minipage}
    
    \caption{(a). The modules highlighted in red represent the three core components of our model: \textbf{MCA}~(Model Class Confusion Analysis Module), \textbf{MCC}~(Multi-Prototype Confused CLIP), and \textbf{FAM}~(Feature Space Alignment Module). (b). The detailed workflow of \textbf{CCF}~(Constructing a Confusion-Aware Feature Center Bank). (c). The specific structure and data flow of \textbf{FAM}~(Feature space Align Module).}
    \label{figure_main}

\end{figure*}

\section{Method}
Let $\mathcal{D}_{s}$=$\{{\mathcal{X}_{s}
,\mathcal{Y}_{s}}\}$=$\{({\mathit{x}^{s}_{i},\mathit{y}^{s}_{i}})\}^{N_s}_{i=1}$ be the unseen source domain dateset, where $N_s$ is the number of source samples, and we define the target domain dataset as
$\mathcal{D}_{t}$=$\{\mathcal{X}_{t}
\}$=$\{\mathit{x}^{t}_{j}\}^{N_t}_{j=1}$, where $N_t$ is the size of the target dateset. In SFDA, the source data cannot be accessed; we could just access the source model $f_{\theta_s}$:$x_s \rightarrow y_s$, which has undergone extensive pre-training on $\mathcal{D}_s$. In general, $f_{\theta_s}$ is composed of a feature extractor $F$ and a classifier $G$ that is usually based on the fully connected layer.

We use the CLIP which consists of an image encoder $\mathcal{I}$ and a text encoder $\mathcal{T}$. During the prediction phase, the CLIP model generates predictions for zero-shot learning by calculating the cosine similarity between text and image features. In our method, during the training process, we freeze the vast majority of parameters in the CLIP model and only fine-tune the model like CoOp~\cite{zhou2022learning} through the text prompt. The text used to compute the text feature can be represented in the following form:"[prefix][cls]". where [cls] is the classname and we use "A picture of a" as the initial prefix.

\begin{algorithm}
\SetKwInOut{Input}{input}
\SetKwInOut{Output}{output}
\caption{Model Class Confusion Process}\label{alg:alg1}

\textbf{Inputs}: Target dataset $\mathcal{D}_t$, Target data $x_i$, source model $f_{\theta_s}$, Multimodal model $Clip$\;

\For{epoch $= 1$ \KwTo $N_e$}{
    \For{$i\leftarrow 0$ \KwTo $\|\mathcal{D}_t\|$}{
        $P^s_i=f_{\theta_s}(x_i)$, $P^c_i=Clip(x_i)$\;
        $w^s_i,w^c_i \leftarrow \textbf{Equation(1)}$\;  
    }
    ${CM^s, CM^c} \leftarrow \textbf{Equation(2)}$\;  
    $\mathcal{M} \leftarrow \textbf{Equation(3) and (4)}$\;  
    $\text{textual prompts} \leftarrow \mathcal{M}$\;  
    $p^c_t \leftarrow \textbf{Equation(6)}$\;  
    $F_c \leftarrow \textbf{Equation(9) and (10)}$\;  
    \textbf{Train}\tcp{Algorithm 2}  
}
\end{algorithm}

\begin{algorithm}
\SetKwInOut{Input}{input}
\SetKwInOut{Output}{output}
\caption{Training Process}\label{alg:alg2}

\textbf{Inputs}: Target dataset $\mathcal{D}_t$, Target data $x_i$, source model $f_{\theta_s}$, Multi-center prediction$p^c_t$ ;

\For{epoch $= 1$ \KwTo $N_e$}{
    \textbf{Pre-operation}\tcp{Algorithm 1} 
    \For{$i\leftarrow 0$ \KwTo $\|\mathcal{D}_t\|$}{
    $p^s_t=f_{\theta_s}(x_i)$\;
    $p^r_t \leftarrow \textbf{Equation(13)}$\;
    Update prompts parameters by minimize $\mathcal{L}_{clip}$\;
    Using feature alignment calculating $F_s$ and $F_c$\;
    Calculating $F_a$ by \textbf{Equation(12)}\;
    Update parameters $\theta_s$ by minimize $\mathcal{L}_s$\;

    }
}
\end{algorithm}

The general architecture of our method can be understood from Fig. \ref{figure_main} (a). According to the three tightly coupled stages: confusion perception, confusion representation,
and confusion alignment, our CGA model mainly comprises three core components: 1. \textbf{Model Class Confusion Analysis Module~(MCA)}: This component quantitatively evaluates the class confusion relationships of the model via a confusion graph, or more specifically, a confusion matrix generated by leveraging the full-sample classification results to of the target dataset across corresponding models; 2. \textbf{Multi-Prototype Confused CLIP~(MCC)}: Based on the class confusion relationships calculated, we employ the multi-prototype prompting strategy to generate the corresponding confusing class texts, then upgrade the CLIP classification method by adding the confusing category text features into the base text features; 3. \textbf{Feature Space Alignment Module~(FAM)}: Building upon the class confusion relationships, this module adaptively weights the target domain features extracted by the source model using classification confidence scores from both the CLIP and the source model to get the center bank of the confusion class features. Then by aligning the projections of target samples calculated by the class feature center banks of the two models, we can implicitly align the feature spaces between two models, thereby transferring CLIP's knowledge at the feature level to optimize the source model.

\subsection{Model Class Confusion Analysis Module}

To estimate potential asymmetrical class confusions in the source model and CLIP, we first propose a method to quantitatively estimate the class confusion relationship matrix. 

We discovered that the class confusion relationships matrix in the model can be inferred by computing class-conditional probability centroids on the target domain. This is because the probabilistic predictions of target domain data inherently encode inter-class confusion patterns, which reflect overlaps and ambiguities between classes. Inspired by SHOT~\cite{liang2020we}, we computed probabilistic centers using a method similar to its approach to compute feature centers. However, our method is not entirely identical to SHOT, which computes feature-based centers; instead, we weight probabilities (rather than features) to get the final probabilistic centers.
In particular, we only assign weights to the most probable candidate classes for each sample, ignoring the classes to which it cannot possibly belong. Consequently,as shown in Eq.\ref{1} we calculate the contribution weights $\boldsymbol{w}_t$ of each sample $x_t$ to all classes.

\begin{equation}
{w}_{t}^{k}=
\begin{cases} 
 P_k(x_t) & \text{if } P_k(x_t) \in \text{top } N \text{ of }P(x_t)  \\
0 & \text{otherwise},
\label{1}
\end{cases}
\end{equation}
where ${P = \delta \circ f_{\theta_s}}$ denotes the probability calculated by the source model $f_{\theta_s}$ and the softmax function $\delta$. $P_k(x_t)$ denotes the $k_{th}$ component of $P(x_t)$, representing the probability that $x_t$ belongs to the $k_{th}$ class. And to eliminate interference, we posit that each probability output contributes exclusively to its $N$ most likely categories. And ${w}_{t}^{k}$ is the result $k_{th}$ of $\boldsymbol{w}_t$. We used the scalar $\boldsymbol{w}_t$ to weight each sample, thus deriving the probability center $\boldsymbol{{CM}_k}$, which represents the center of the class for the class $k_{th}$, and the matrix $\boldsymbol{CM}$ could be interpreted as the confusion class relationship matrix.

\begin{equation}
\boldsymbol{{CM}_k} = \frac{\Sigma_{x_t\in\mathcal{X}_{t}}{w_t^k} P(x_t)}{\Sigma_{x_t\in\mathcal{X}_{t}}{w_t^k}}
\label{2}
\end{equation}

Considering that if the $\boldsymbol{{CM}_{ki}}$~($i\neq k$)is bigger than $\frac{1}{C}$~($C$ is the number of classes), then we can infer that there is a certain degree of confusion between class $k$ and class $i$.

So, from the matrix $\boldsymbol{CM}$, we generate ordered class confusion pairs based on elements in the matrix that exceed $\frac{1}{C}$. For example, if $\boldsymbol{{CM}_{ki}}>\frac{1}{C}$,  we create a confusion pair $<i,j>$ where $i$ is the primary class and $j$ is the secondary class. 
Subsequently, we compile all the asymmetrical $<i,j>(i\neq j)$ obtained into a collective set $\mathcal{M}_{pre}$:
\begin{equation}
\mathcal{M}_{sim}=\{<i,j>\mid{\boldsymbol{{CM}_{ij}}\geq T_{sim} \enspace and\; i\neq j\}}
\label{3}
\end{equation}
where $T_{sim}=max(\frac{1}{C},K_C)$ serves to constrain the size of the set $\mathcal{M}_{sim}$, with $K_C$ denoting the $C_{th}$ largest value among all $CM_{ij}(i\neq j)$ entries.

In addition, we add all pairs$<i,j>$ with $i=j$ to $\mathcal{M}_{sim}$ to facilitate subsequent operations.

\begin{equation}
\mathcal{M}=\mathcal{M}_{sim}\cup\{{<i,i>}\mid i=1,2,...,C\}
\label{4}
\end{equation}

Based on the aforementioned computational approach, we can obtain the class confusion relation matrices $\boldsymbol{CM^{c}}$ and $\boldsymbol{CM^{s}}$ for the CLIP and the source model, along with the ordered confusion class pair sets $\mathcal{M}^{c}$ and $\mathcal{M}^{s}$.

\subsection{Multi-Center CLIP}
After getting the confusion pairs set $\mathcal{M}^s$, we aim to customize CLIP with it to get the soft predictions that could mitigate the asymmetrical class confusion patterns observed in the source model. Therefore, we aim to achieve our goal by modifying the text input side of CLIP.

To match the asymmetrical class confusion, for each element in $\mathcal{M}^s$, we can generate the corresponding class texts. 
In CLIP, during zero-shot classification of images, the image features $F^I=\mathcal{I}(x_t)$ are first calculated, then the similarity is calculated with the text features $F^T$ generated from the given category descriptions using cosine similarity, generating the classification results. However, in our approach, we use $\mathcal{M}^s$ to generate class labels, which also include confused class labels. For instance, for $<i,i>$ in $\mathcal{M}^s$, we generate the class text "$class(i)$", and for $<i,j>$ where 
$i\neq j$, we create the confused  class text "$class(i)$ looks like a $class(j)$", where $class$ is the function which can get the corresponding class text. 

Following this, we categorize the acquired features, clustering text features that share the same principal category into the same text feature subsets. 
\begin{equation}
\boldsymbol{\mathcal{F}^T_i}=\mathcal{T}(\mathcal{G}(\{<i,j>\mid <i,j> \in \mathcal{M}^s \enspace and \enspace j =1,2,...,C\}))
\label{5}
\end{equation}
where $\mathcal{G}$ is the function that translates the similarity relationships within the set $\mathcal{M}$ into text form. 

We then compute the similarity with the image features for each element in the subset and select the highest similarity score as the logits for that class. The final classification result ${\boldsymbol{p^c_t}}$ is achieved by subsequently applying the softmax function to these logits. 
\begin{equation}
 \begin{aligned}
        & logit_i=max({cos(F^I,\boldsymbol{\mathcal{F}^T_i})})\\
        &\boldsymbol{p^c_t} = \delta(logit_1,logit_2,...,logit_C)
\end{aligned}
 \label{6}
\end{equation}
where $cos$ is the function of the cosine similarity calculation function. And we consider $\boldsymbol{p^c_t}$ to be a classification result with finer-grained discriminative power compared to the original output of CLIP, so reducing the discrepancy between $\boldsymbol{p^c_t}$  and the output of the source model could to some extent alleviate the confusion relationship in the source model so that we can extract knowledge from CLIP in a more targeted manner. 

To summarize the method described above, we define the process of computing $\boldsymbol{p^c_t}$ from input target data $x_t$ as $\boldsymbol{p^c_t}=P^{\prime}(x_t)$

We noticed that the CLIP may yield incorrect classification results for the confusion classes text features. Furthermore, for the degree of similarity between the primary and secondary classes within confused classes, we need to provide quantitative guidance for CLIP. For instance, with the confusion class "a class[i] looks like a class[j]", we still require fine-grained adjustments to the specific similarity between the two classes. Therefore, we calculate the similarity of the confusion classes based on the result of Eq.\ref{2}.
For each $<i,j>\in \mathcal{M}$ ($i\neq j$), we could first calculate the degree of similarity between the $i_{th}$ class and the
$j_{th}$ through the ratio $\mathcal{R}_{sim(i,j)}$ between ${\boldsymbol{CM}}_{ii}$ and ${\boldsymbol{CM}}_{ij}$. 
\begin{equation}
\mathcal{R}_{sim(i,j)} =\frac{\boldsymbol{CM}_{ii}}{\boldsymbol{CM}_{ij}}
\label{7}
\end{equation}

Subsequently, we use $\mathcal{R}_{sim(i,j)}$ to estimate the probabilistic centroid $\overline{p}_{ij}^{sim}$ associated with the confusion pair (i,j) that satisfies two conditions: (1) It has a value of zero at all positions except the positions of $i_{th}$ and $j_{th}$. (2) The sum of the values at the
$i_{th}$ and the $j_{th}$ positions is equal to one, with their ratio being $\mathcal{R}_{sim(i,j)}$. For the non-confusion class $<i,i> \in \mathcal{M}$, we construct a probability vector that is one at the $i_{th}$ position and zero at all other positions.

Then, we achieve fine-grained adjustment of the confusion class features within CLIP by minimizing the following loss $\mathcal{L}_{r}$:
\begin{equation}
    \mathcal{L}_{r} = KL( \delta(cos(F^T_{ij},\boldsymbol{\mathcal{F}^T})),\overline{p}_{ij}^{sim})
    \label{8}
\end{equation}

First, we utilize the text features of the confusion classes $F^T_{ij}=\mathcal{T}(\mathcal{G}(<i,j>))$ in place of the image features, and classify them using the base non-confusion class text features $\boldsymbol{\mathcal{F}^T}=\mathcal{T}(\mathcal{G}(\{<i,i>\mid i=1,2,...,C\}$ to obtain the classification results for the confusion classes. 
Then, we minimize the Kullback-Leibler(KL) divergence between the softmax output of the classification results and $\overline{p}_{ij}$, achieving accurate and fine-grained adjustments for each distinct pair of confusion classes.
\subsection{Feature Space Alignment Module} 
At the start of each epoch, we compute the class feature centers for both models throughout the target dataset as shown in Fig. \ref{figure_main} (b). In particular, the classes requiring feature center computation are obtained from the set $\mathcal{M}$, including the confusion classes. However, due to significant representational discrepancies between CLIP's image features and the source model's feature space, directly aligning the respective features proves infeasible. To address this, we leverage the features calculated by source model on target data as the common basis, while utilizing classification outcomes from both models to guide the construction of corresponding class centers, respectively.

Unlike the class probability computation in Eq.\ref{2}, here our focus shifts to deriving precise feature centers. So we use a new method to calculate the center of features inspired by BMD~\cite{qu2022bmd}. 
We need to calculate a score function $S_{ij}$ based on the classification probabilities of each sample to facilitate the classification to find the center of the features of $<i,j>$ in $\mathcal{M}$. 
\begin{equation}
\begin{aligned}
    {R}(\boldsymbol{p})&=\frac{\boldsymbol{p}_{i}}{\boldsymbol{p}_{j}} \\
    S_{ij}(\boldsymbol{p}) &= \frac{1}{2}(\boldsymbol{p_i}+\boldsymbol{p_j})-\frac{1}{2}|\frac{(R(\boldsymbol{p})-\mathcal{R}_{sim(i,j)})}{(R(\boldsymbol{p})+\mathcal{R}_{sim(i,j)})}| \\
\end{aligned}    
 \label{9}
\end{equation}



\textbf{Remark}: The score function $S_{ij}$ comprises two components. The first part calculates the sum of $p_i$ and $p_j$. 
And the second part computes the distance between the ratio of sample probabilities $R(p)$ and the corresponding ratio $\mathcal{R}_{sim(i,j)}$ derived from the confusion matrix ${\boldsymbol{CM}}$, followed by normalization to ensure a symmetric distribution of the ratios around $\mathcal{R}_{sim(i,j)}$.
To elucidate the specific meanings of these two components, the first term ensures that higher predicted probabilities are more likely to belong to either class $i$ or class $j$ , rather than any other categories; the second term guarantees that the calculated metric yields a larger value when the ratio between the probabilities of class $i$ and class $j$ is closer to the target ratio corresponding to our estimated confusion center. Furthermore, since both components are normalized to the range [0,1], they can be directly summed with equal weighting to form the final composite metric.
Additionally, when $i=j$, the function $S(p)$ degenerates into $p_i$, meaning that for the non-confusing classes, the probability of the corresponding class is directly taken as the result.

Then we select the features of the top$-M$ scoring samples for each class and compute the center of the features using a uniform weight. 

\begin{equation}
\begin{aligned}
\mathcal{M}_{ij}&=\mathop{argmax}\limits_{|\mathcal{M}_{ij}|=M \atop x_t \in \mathcal{X}_{t} }  S_{ij}(f_\theta(x_t)) \\
    F_c &= \frac{1}{M} \sum \limits_{k \in \mathcal{M}_{ij}}F(x_k)
    \label{10}
\end{aligned}
\end{equation}
 
After getting the center banks, as shown in Fig. \ref{figure_main} (c), during the training phase, we treat each feature center bank as a set of basis vectors that structurally approximate the model's feature space on the target dataset. Subsequently, using MLP projection mechanisms, we project the input features $F_t$ into the corresponding space by computing a weighted combination of the centers in the feature bank, then the projected features are fused with the original features through residual summation within the residual network framework, producing two distinct outputs $F_{clip}$ and $F_s$.

Finally, contrastive learning is used to minimize the distance between $F_c$ and $F_s$, achieving implicit alignment between CLIP and the source model.
 \begin{equation}
     \mathcal{L}_{ct} = Simclr(\boldsymbol{F_s},\boldsymbol{F_{clip}})
     \label{11}
 \end{equation}
  Simclr~\cite{chen2020simple} is a self-supervised contrastive loss function.
 
The augmented feature $F_a$ is obtained by dynamically weighting the $F_{clip}$ and $F_s$. The weight $\mathcal{W}$ is a simple MLP that is connected only to the input feature $F_t$.
 \begin{equation}
     F_a = \mathcal{W}(F_t)F_s+(1-\mathcal{W}(F_t))F_{clip}
     \label{12}
 \end{equation}

\subsection{Loss Function} 
To extract more accurate information from 
$\boldsymbol{p^c_t}$ and 
$\boldsymbol{p_t^s}=\delta(f_\theta(x))$, we refine the probabilities of these two distributions into a single consolidated probability $\boldsymbol{p^r_t}$ and subsequently remove its gradient.
\begin{equation}
 \boldsymbol{p^r_t}=\left\{
\begin{aligned}
\mathop{argmin}\limits_{H(\boldsymbol{p_t})}(\boldsymbol{p^c_t},\boldsymbol{p_t^s})\quad &, if  A(\boldsymbol{p^c_t},\boldsymbol{p_t^s}) \\
\omega\boldsymbol{p^c_t}+
(1-\omega)\boldsymbol{p_t^s}\quad &, otherwise
\end{aligned}
\right.
\label{13}
\end{equation}
where $H(x) = -\sum_{i=1}^{n} p(x_i) \log p(x_i)$ calculates the information entropy of probability $x$ and the condition is fulfilled $A(x_1,x_2)$ if and only if $argmax(x_1)=argmax(x_2)$ is satisfied. 

If the pseudo-labels of $\boldsymbol{p_t^s}$ and $\boldsymbol{p^c_t}$ are consistent, then $\boldsymbol{p^r_t}$ is the one with the lower information entropy between them; otherwise, we use the information entropies of both to perform weighting. 
\begin{equation}
\omega = \frac{H(\boldsymbol{p_t^s})}{H(\boldsymbol{p_t^s})+H(\boldsymbol{p^c_t})}
\label{14}
\end{equation}

After getting $\boldsymbol{p^r_t}$ in Eq.\ref{13}, we then adopt $\boldsymbol{p^r_t}$ as the reference distribution to optimize the source model, the FAM, and CLIP’s prompts by minimizing the KL divergence between their respective outputs:
\begin{equation}
\begin{aligned}
    \mathcal{L}_{c} = &KL(\boldsymbol{{p}^c_t},\boldsymbol{{p}^r_t}) \\
    \mathcal{L}_{s} = &KL(\boldsymbol{{p}^s_t},\boldsymbol{{p}^r_t}) \\
    \mathcal{L}_{a} =& KL(\boldsymbol{\delta (G(F_a))},\boldsymbol{{p}^r_t}) \\
\end{aligned}    
\label{15}
\end{equation}

The loss function of CLIP $\mathcal{L}_{clip}$ consists of the downstream task alignment loss $\mathcal{L}_{c}$ and the text refinement loss $\mathcal{L}_{r}$ shown below:
\begin{equation}
    \mathcal{L}_{clip}  = \mathcal{L}_{r}+\mathcal{L}_{c}
    \label{16}
\end{equation}

The loss $\mathcal{L}_{source}$ of the source model and \textbf{FAM} consists of $\mathcal{L}_{s}$ to align $\boldsymbol{{p}^s_t}$ and $\boldsymbol{{p}^r_t}$, optimizing the source model using CLIP's knowledge, $\mathcal{L}_{a}$ for optimizing the \textbf{FAM} and $\mathcal{L}_{ct}$ to implicitly align the feature spaces of the source model and CLIP.
\begin{equation}
    \mathcal{L}_{source}  = \mathcal{L}_{s}+\alpha\mathcal{L}_{a}+\gamma\mathcal{L}_{ct}
    \label{17}
\end{equation}
where $\alpha$, $\beta$ and $\gamma$ are the hyper-parameters.

In general, our method consists of two processes: a pre-processing process at the beginning of each epoch and the training process, which are detailed in \textbf{Algorithm 1} and \textbf{Algorithm 2}, respectively. All operations in \textbf{Algorithm 1} are recomputed at the beginning of each epoch to dynamically adapt to the evolving source model during training.
Upon completion of training, we exclusively employ the source model (typically ResNet) for downstream tasks or testing, while discarding both CLIP and the FAM network as they solely serve auxiliary roles during the training phase.

\begin{table*}[t]
\centering
\caption{Accuracy (\%) on VisDA for UDA and source-free UDA methods (ResNet-101). }
 \setlength{\tabcolsep}{4pt}
\begin{tabular}{c c c|c c c c c c c c c c c c|c}
\hline
Method & SF & CLIP &plane & bcycl& bus& car& horse& knife &mcycl& person &plant &sktbrd &train &truck &Avg. \\
\hline
DAPL-RN~\cite{ge2023domain} & \usym{2717}&\usym{2714}&97.8&83.1&88.8&77.9&97.4&91.5&94.2&79.7&88.6&89.3&92.5&62.0&86.9\\
PADCLIP-RN~\cite{lai2023padclip}&\usym{2717} &\usym{2714}&96.7&88.8&87.0&82.8&97.1&93.0&91.3&83.0&95.5&91.8&91.5&63.0&88.5\\
ADCLIP-RN~\cite{singha2023ad} &\usym{2717}&\usym{2714}& 98.1&83.6&91.2&76.6&98.1&93.4&96.0&81.4&86.4&91.5&92.1&64.2&87.7\\
PDA-R~\cite{bai2024prompt}&\usym{2717}&\usym{2714}&97.2&82.3&89.4&76.0&97.4&87.5&95.8&79.6&87.2&89.0&93.3&62.1&86.4\\
DAMP-R~\cite{du2024domain}&\usym{2717}&\usym{2714} &97.3&91.6&89.1&76.4&97.5&94.0&92.3&84.5&91.2&88.1&91.2&67.0&88.4\\
\hline
SHOT~\cite{liang2020we}& \usym{2714} &\usym{2717} &95.0&87.4&80.9&57.6&93.9&94.1&79.4&80.4&90.9&89.8&85.8&57.5&82.7\\
NRC~\cite{yang2021exploiting}& \usym{2714} &\usym{2717} &96.8&91.3&82.4&62.4&96.2&95.9&86.1&90.7&94.8&94.1&90.4&59.7&85.9\\
GKD~\cite{tangmodel}& \usym{2714} &\usym{2717} &95.3&87.6&81.7&58.1&93.9&94.0&80.0&80.0&91.2&91.0&86.9&56.1&83.0\\
AaD~\cite{yang2022attracting}& \usym{2714} &\usym{2717} &97.4&90.5&80.8&76.2&97.3&96.1&89.8&82.9&95.5&93.0&92.0&64.7&88.0\\
AdaCon~\cite{chen2022contrastive}& \usym{2714} &\usym{2717} &97.0&84.7&84.0&77.3&96.7&93.8&91.9&84.8&94.3&93.1&94.1&49.7&86.8\\
CoWA~\cite{lee2022confidence}& \usym{2714} &\usym{2717} &96.2&89.7&83.9&73.8&96.4&97.4&89.3&86.8&94.6&92.1&88.7&53.8&86.9\\
ELR~\cite{yi2023source}& \usym{2714} &\usym{2717} &97.1&89.7&82.7&62.0&96.2&97.0&87.6&81.2&93.7&94.1&90.2&58.6&85.8\\
PLUE~\cite{litrico2023guiding}& \usym{2714} &\usym{2717} &94.4&91.7&89.0&70.5&96.6&94.9&92.2&88.8&92.9&95.3&91.4&61.6&88.3\\
CPD~\cite{zhou2024source}& \usym{2714} &\usym{2717} &96.7&88.5&79.6&69.0&95.9&96.3&87.3&83.3&94.4&92.9&87.0&58.7&85.5\\
TPDS~\cite{tang2024source1}& \usym{2714} &\usym{2717} &97.6&91.5&89.7&83.4&97.5&96.3&92.2&82.4&96.0&94.1&90.9&40.4&87.6\\
\hline

DIFO-V~\cite{tang2024source}&\usym{2714}&\usym{2714}&97.5& 89.0 &\textbf{90.8} &83.5 &97.8 &97.3& 93.2& 83.5 &95.2 &\textbf{96.8}& 93.7 &{65.9} &90.3\\
ProDe-V~\cite{tang2024proxy}&\usym{2714}&\usym{2714}&98.3&\textbf{92.0}&87.3&84.4&98.5&97.5&94.0&86.4&95.0&96.1&94.2&\textbf{75.6}&91.6\\
CGA(ours)&\usym{2714}&\usym{2714}&\textbf{98.9}&91.6&87.5&\textbf{88.2}&\textbf{99.2}&\textbf{98.3}&\textbf{94.9}&\textbf{87.5}&\textbf{96.0}&95.0&\textbf{94.4}&74.5&\textbf{92.2}\\
\hline

\end{tabular}
\label{table_visda}
\end{table*}

\begin{table*}[t]
\centering
\caption{Accuracy (\%) on Office-Home for UDA and source-free UDA methods (ResNet-50).}
 \setlength{\tabcolsep}{4pt}
\begin{tabular}{c c c|c c c c c c c c c c c c|c}
\hline
Method & SF & CLIP& A→C&A→P &A→R& C→A& C→P &C→R& P→A& P→C& P→R& R→A& R→C &R→P& Avg.\\
\hline
DAPL-RN~\cite{ge2023domain}&\usym{2717}&\usym{2714} &54.1&84.3&84.8&74.4&83.7&85.0&74.5&54.6&84.8&75.2&54.7&83.8&74.5\\
PADCLIP-RN~\cite{lai2023padclip}&\usym{2717}&\usym{2714} &57.5&84.0&83.8&77.8&85.5&84.7&76.3&59.2&85.4&78.1&60.2&86.7&76.6\\
ADCLIP-RN~\cite{singha2023ad}&\usym{2717}&\usym{2714} &55.4&85.2&85.6&76.1&85.8&86.2&76.7&56.1&85.4&76.8&56.1&85.5&75.9\\
PDA-R~\cite{bai2024prompt}&\usym{2717}&\usym{2714}&55.4&85.1&85.8&75.2&85.2&85.2&74.2&55.2&85.8&74.7&55.8&86.3&75.3\\
DAMP-R~\cite{du2024domain}&\usym{2717}&\usym{2714} &59.7&88.5&86.8&76.6&88.9&87.0&76.3&59.6&87.1&77.0&61.0&89.9&78.2\\

\hline

SHOT~\cite{liang2020we}& \usym{2714} &\usym{2717}&56.7&77.9&80.6&68.0&78.0&79.4&67.9&54.5&82.3&74.2&58.6&84.5&71.9\\
NRC~\cite{yang2021exploiting}& \usym{2714} &\usym{2717}&57.7&80.3&82.0&68.1&79.8&78.6&65.3&56.4&83.0&71.0&58.6&85.6&72.2\\
GKD~\cite{tangmodel}& \usym{2714} &\usym{2717}&56.5&78.2&81.8&68.7&78.9&79.1&67.6&54.8&82.6&74.4&58.5&84.8&72.2\\

AaD~\cite{yang2022attracting}& \usym{2714} &\usym{2717}&59.3&79.3&82.1&68.9&79.8&79.5&67.2&57.4&83.1&72.1&58.5&85.4&72.7\\
AdaCon~\cite{chen2022contrastive}& \usym{2714} &\usym{2717}&47.2&75.1&75.5&60.7&73.3&73.2&60.2&45.2&76.6&65.6&48.3&79.1&65.0\\
CoWA~\cite{lee2022confidence}& \usym{2714} &\usym{2717}&56.9&78.4&81.0&69.1&80.0&79.9&67.7&57.2&82.4&72.8&60.5&84.5&72.5\\
ELR~\cite{yi2023source}& \usym{2714} &\usym{2717}&58.4&78.7&81.5&69.2&79.5&79.3&66.3&58.0&82.6&73.4&59.8&85.1&72.6\\
PLUE~\cite{litrico2023guiding}& \usym{2714} &\usym{2717}&49.1&73.5&78.2&62.9&73.5&74.5&62.2&48.3&78.6&68.6&51.8&81.5&66.9\\
CPD~\cite{zhou2024source}& \usym{2714} &\usym{2717}&59.1&79.0&82.4&68.5&79.7&79.5&67.9&57.9&82.8&73.8&61.2&84.6&73.0\\
TPDS~\cite{tang2024source1}& \usym{2714} &\usym{2717}&59.3&80.3&82.1&70.6&79.4&80.9&69.8&56.8&82.1&74.5&61.2&85.3&73.5\\
\hline
DIFO-V ~\cite{tang2024source}&\usym{2714} &\usym{2714} &70.6&90.6&88.8&82.5&90.6&88.8&80.9&70.1&88.9&83.4&70.5&91.2&83.1\\
ProDe-V ~\cite{tang2024proxy}&\usym{2714} &\usym{2714}&\textbf{74.6}&\textbf{92.9}&92.4&84.4&93.0&92.2&83.8&\textbf{74.8}&\textbf{92.4}&84.9&\textbf{75.2}&\textbf{93.7}&\textbf{86.2}\\
CGA(ours)&\usym{2714}&\usym{2714}&74.1&92.8&\textbf{92.6}&\textbf{86.0}&\textbf{93.1}&\textbf{92.4}&\textbf{86.0}&73.3&92.2&\textbf{86.0}&73.0&93.6&\textbf{86.3}

\\
\hline
\end{tabular}
\label{table_office-home}
\end{table*}

\begin{table*}[t]
\centering
\caption{Accuracy (\%) on DomainNet-126 for UDA and source-free UDA methods (ResNet-50). }
 \setlength{\tabcolsep}{5pt}
\begin{tabular}{c c c|c c c c c c c c c c c c|c}

\hline
Method & SF & CLIP &C→P & C→R& C→S& P→C& P→R& P→S &R→C& R→P &R→S &S→C &S→P &S→R &Avg. \\
\hline
DAPL-RN~\cite{ge2023domain}&\usym{2717}&\usym{2714} &72.4&87.6&65.9&72.7&87.6&65.6&73.2&72.4&66.2&73.8&72.9&87.8&74.8\\

ADCLIP-RN~\cite{singha2023ad}&\usym{2717}&\usym{2714} &71.7&88.1&66.0&73.2&86.9&65.2&73.6&73.0&68.4&72.3&74.2&89.3&75.2\\

DAMP-R~\cite{du2024domain}&\usym{2717}&\usym{2714} &76.7&88.5&71.7&74.2&88.7&70.8&74.4&75.7&70.5&74.9&76.1&88.2&77.5\\

\hline

SHOT~\cite{liang2020we}& \usym{2714} &\usym{2717}&63.5&78.2&59.5&67.9&81.3&61.7&67.7&67.6&57.8&70.2&64.0&78.0&68.1\\
NRC~\cite{yang2021exploiting}& \usym{2714} &\usym{2717}&62.6&77.1&58.3&62.9&81.3&60.7&64.7&69.4&58.7&69.4&65.8&78.7&67.5\\
GKD~\cite{tangmodel}& \usym{2714} &\usym{2717}&61.4&77.4&60.3&69.6&81.4&63.2&68.3&68.4&59.5&71.5&65.2&77.6&68.7\\

AdaCon~\cite{chen2022contrastive}& \usym{2714} &\usym{2717}&60.8&74.8&55.9&62.2&78.3&58.2&63.1&68.1&55.6&67.1&66.0&75.4&65.4\\
CoWA~\cite{lee2022confidence}& \usym{2714} &\usym{2717}&64.6&80.6&60.6&66.2&79.8&60.8&69.0&67.2&60.0&69.0&65.8&79.9&68.6\\

PLUE~\cite{litrico2023guiding}& \usym{2714} &\usym{2717}&59.8&74.0&56.0&61.6&78.5&57.9&61.6&65.9&53.8&67.5&64.3&76.0&64.7\\
TPDS~\cite{tang2024source1}& \usym{2714} &\usym{2717}&62.9&77.1&59.8&65.6&79.0&61.5&66.4&67.0&58.2&68.6&64.3&75.3&67.1\\
\hline
DIFO-V ~\cite{tang2024source}&\usym{2714} &\usym{2714} &76.6&87.2&74.9&80.0&87.4&75.6&80.8&77.3&75.5&80.5&76.7&87.3&80.0\\
ProDe-V ~\cite{tang2024proxy}&\usym{2714} &\usym{2714}&83.2&\textbf{92.4}&79.0&85.0&\textbf{92.3}&79.3&85.5&83.1&79.1&85.5&83.4&\textbf{92.4}&85.0\\
CGA(ours)&\usym{2714}&\usym{2714}&\textbf{85.5}&90.5&\textbf{83.2}&\textbf{85.1}&91.1&\textbf{84.2}&\textbf{87.0}&\textbf{86.0}&\textbf{83.7}&\textbf{86.2}&\textbf{84.9}&91.5&\textbf{86.6}

\\
\hline
\end{tabular}
\label{table_domainnet-126}
\end{table*}

\begin{table}[t]
\centering
\caption{Accuracy (\%) on Office-31 for UDA and source-free UDA methods (ResNet-50).}
\setlength{\tabcolsep}{0.3mm}
\begin{tabular}{c|c c|c c c c c c|c}
\hline

Method&SF&C&A$\rightarrow$D&A$\rightarrow$ W&D$\rightarrow$A&D$\rightarrow$W&W$\rightarrow$A&W$\rightarrow$D& AVG.\\
\hline

Source&\usym{2717}&\usym{2717}&79.1&76.9&59.9&95.5&61.4&98.8&78.6\\
\hline

SHOT~\cite{liang2020we}&\usym{2714}&\usym{2717}&93.7&91.1&74.2&98.2&74.6&100.&88.6\\

NRC~\cite{yang2021exploiting}&\usym{2714}&\usym{2717}&96.0&90.8&75.3&99.0&75.0&100.&89.4\\



AaD~\cite{yang2022attracting}&\usym{2714}&\usym{2717}&96.4&92.1&75.0&99.1&76.5&100.&89.9\\

AdaCon~\cite{chen2022contrastive}&\usym{2714}&\usym{2717}&87.7&83.1&73.7&91.3&77.6&72.8&81.0\\

CoWA~\cite{lee2022confidence}&\usym{2714}&\usym{2717}&94.4&95.2&76.2&98.5&77.6&99.8&90.3\\

ELR~\cite{yi2023source}&\usym{2714}&\usym{2717}&93.8&93.3&76.2&98.0&76.9&100.&89.6\\

PLUE~\cite{litrico2023guiding}&\usym{2714}&\usym{2717}&89.2&88.4&72.8&97.1&69.6&97.9&85.8\\


TPDS~\cite{tang2024source1}&\usym{2714}&\usym{2717}&97.1&94.5&75.7&98.7&75.5&99.8&90.2\\
\hline
DIFO-V~\cite{tang2024source}&\usym{2714}&\usym{2714}&97.2&95.5&83.0&97.2&\textbf{83.2}&98.8&92.5\\

ProDe-V~\cite{tang2024proxy}&\usym{2714}&\usym{2714}&96.6&\textbf{96.4}&\textbf{83.1}&96.9&82.9&99.8&92.6\\
CGA(ours)&\usym{2714}&\usym{2714}&\textbf{97.4}&95.7&82.5&\textbf{98.5}&82.0&\textbf{100.}&\textbf{92.7} \\
\hline

\end{tabular}
\label{table_office-31}
\end{table}

\section{Experiment}
\subsection{Implementation details}
To evaluate the effectiveness of the proposed method, we employ experiments on some challenging datasets, including the small-scaled datasets of \textbf{Office-31}~\cite{li2017deeper} and \textbf{Office-Home}~\cite{venkateswara2017deep}, and large-scaled datasets of \textbf{VisDA}~\cite{peng2017visda} and \textbf{DmainNet-126}~\cite{saito2019semi}. 

We employ ResNet~\cite{he2016deep} as the foundational architecture for our model, which consists mainly of a feature extractor and a classification layer. To ensure a fair comparison, we utilize the same ResNet variant as our counterpart for the backbone of our model. Specifically, for Office-31, Office-Home and DomainNet-126, we have chosen ResNet-50 to serve as the backbone. In VisDA, we used ResNet-101. For CLIP, we uniformly use ViT-B/16 as the backbone of the image encoder.


We used identical hyperparameter settings in three datasets: parameter $\alpha$ was set to 0.5 for good results, while $\gamma$ was fixed at 0.05 and its sensitivity analysis is presented in Fig. \ref{figure_parameter}(b).

For $N$ in the computation of the confusion matrix in Eq.\ref{1}, we set it to 2 in all settings. And the analysis compared to other settings can be seen in Fig. \ref{figure_parameter}(a).

\subsection{Comparison Performance}
We conducted tests of our method and several comparative approaches in three datasets, and the results are documented in Tables I-IV. In each column, the bold figures indicate the highest accuracy achieved in the respective tasks, and SF and CLIP~(or C) indicate whether the method is a source-free setting and whether the CLIP model has been used, respectively.

Tables I-IV demonstrate that our method outperforms the previously best methods on average in three datasets.  Specifically, of the 12 tasks on Office-Home, 6 achieved optimal results, and of the 12 categories on VisDA, 8 obtained the best performance, with half the configurations on Office-31 being optimal, and in DomainNet-126, 9 settings of 12 settings achieved the best. In particular, on the VisDA dataset, among a group of semantically similar classes: car, bus, train, and truck, we observed an  improvement of 3.9\% over the previous best method. These results indicate that our CGA method improves cross-domain performance and showcases its fine-grained classification capability for confusable classes. Additionally, we note that while our method achieves modest average improvements on small-scale datasets (Office-31 and Office-Home), it demonstrates significantly more pronounced gains on larger-scale datasets (VisDA and DomainNet-126). This indicates that our approach is particularly well suited for handling large-scale data scenarios.

\begin{table}[t]
\centering
\caption{Ablation accuracy (\%) on Office-31,Office-Home and VisDA.}
\renewcommand{\arraystretch}{1.1}
\label{table5}
\begin{tabular}{c c c|c c c |c}
\hline
$\mathcal{L}_{c}$&$\mathcal{L}_{ct}$&$\mathcal{L}_{r}$&Office-31&Office-Home&VisDA&AVG\\
\hline
\usym{2714}&\usym{2714}&\usym{2714}&\textbf{92.7}&\textbf{86.2}&\textbf{92.2}&\textbf{90.3}\\
\usym{2717}&\usym{2714}&\usym{2714}&90.9&82.2&90.5&87.9\\
\usym{2714}&\usym{2717}&\usym{2714}&91.0&83.0&91.3&88.4\\
\usym{2714}&\usym{2714}&\usym{2717}&91.9&83.6&91.8&89.1\\
\hline
\multicolumn{3}{c|}{\textbf{CGA} w/o MCC}&87.5&84.8&91.4&87.9\\

\hline
\end{tabular}

\label{table_ablation}
\end{table}

\section{Discussion}
\subsection{Ablation study.}
In Table \ref{table_ablation}, we report the results of the ablation study for various settings in 3 different datasets.
We conducted a quantitative analysis of the functional roles of individual modules by removing specific loss components and related methods in the experiment. When we remove $\mathcal{L}_c$, we observe a decrease in the overall performance of the model. Similarly, when we remove $\mathcal{L}_r$, we observe that it results in only a minor performance drop compared to removing $\mathcal{L}_c$. We hypothesize that this occurs because $\mathcal{L}_c$ enables CLIP to partially acquire relevant knowledge from the source model, thus mitigating the negative impact caused by the lack of Lr of losing qualitative and quantitative constraints on confusing text classes. When $\mathcal{L}_{ct}$ is removed, the model's mean Average Precision drops by 2$\%$, demonstrating the effectiveness of the feature-space alignment module.
Finally, we evaluated the impact of removing MCC~(Multi-Prototype CLIP), directly using standard CLIP classification results in our experiments. We found that its impact on the entire experiment was substantial and only achieved a marginal improvement compared to the results of the direct zero-shot classification using CLIP. This demonstrates the effectiveness of MCC in addressing issues caused by class confusion.
\begin{figure}[t]
\begin{minipage}{\linewidth}
    \centering
    \includegraphics[scale=0.35]{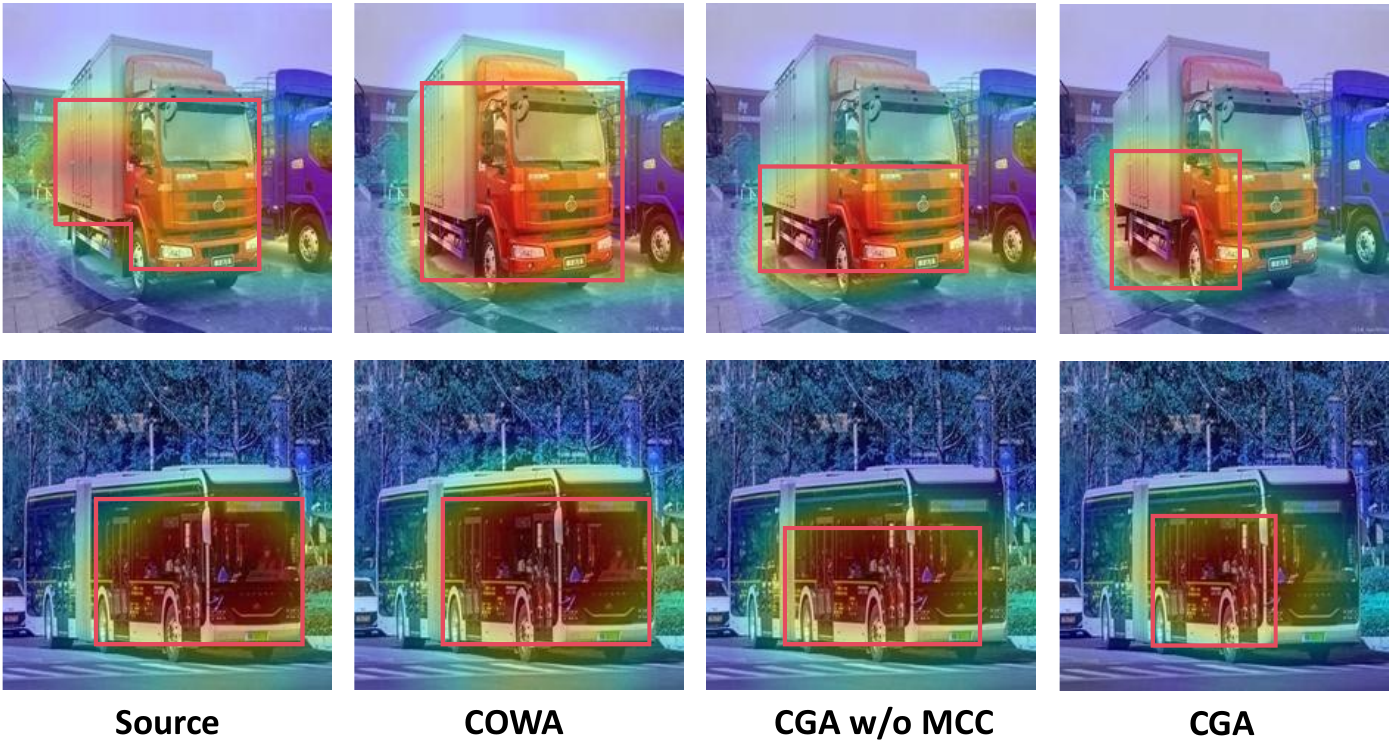}
    \end{minipage}
    
    \caption{The Grad-CAM visualizations of Source, COWA, CGA w/o MCC and CGA models trained in Visda.}
    \label{figure_attension}

\end{figure}

\begin{table}[t]
\centering
\renewcommand{\arraystretch}{1.1}
\caption{Accuracy(\%) of CLIP and its fine-tuned variant models in our method.}
\setlength{\tabcolsep}{0.3mm}
\label{table4}
\begin{tabular}{c |c c c c |c c c c c |c}
\hline
{Method} & \multicolumn{4}{c|}{Office-31} & \multicolumn{5}{c|}{Office-Home} & VisDA\\
&→A&→W&→D&Avg.&→A&→C&→P&→R&Avg.&Sy→Re\\
\hline
CLIP&78.1&79.6&78.9&78.9&81.9&62.9&90.1&88.5&80.8&86.9\\
MCC&75.3&79.6&76.3&77.1&77.7&56.6&82.5&81.0&75.4&74.7\\
MCC w $\mathcal{L}_r$&78.9&81.3&83.1&81.1&81.6&65.5&87.8&88.5&80.9&87.7\\
MCC in CGA&\textbf{82.1}&\textbf{95.5 }&\textbf{94.2}&\textbf{90.6}&\textbf{84.1}&\textbf{69.2}&\textbf{91.2}&\textbf{90.5}&\textbf{83.8}&\textbf{92.2}\\
\hline
\end{tabular}

\label{table_ablation-clip}
\end{table}

\begin{figure}[t]
\begin{minipage}{\linewidth}
    \centering
    \includegraphics[scale=0.5]{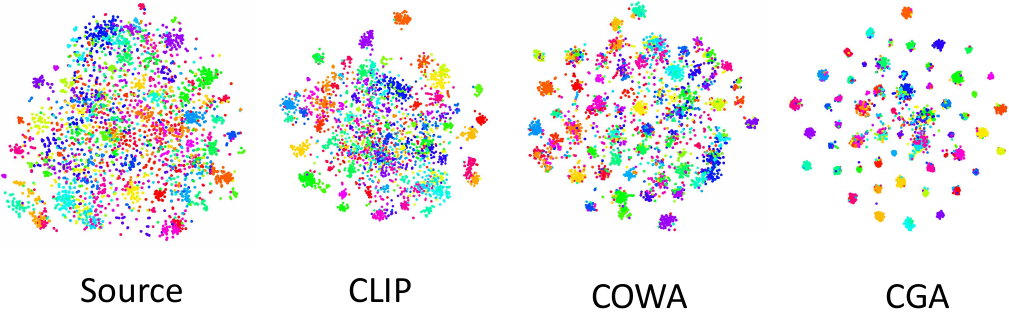}
    \end{minipage}
    
    \caption{Feature distribution visualization comparison on transfer task Ar→Cl in Office-Home by t-SNE.}
    \label{figure_tsne}

\end{figure}

\begin{figure}[t]
\begin{minipage}{\linewidth}
    \centering
    \includegraphics[scale=0.4]{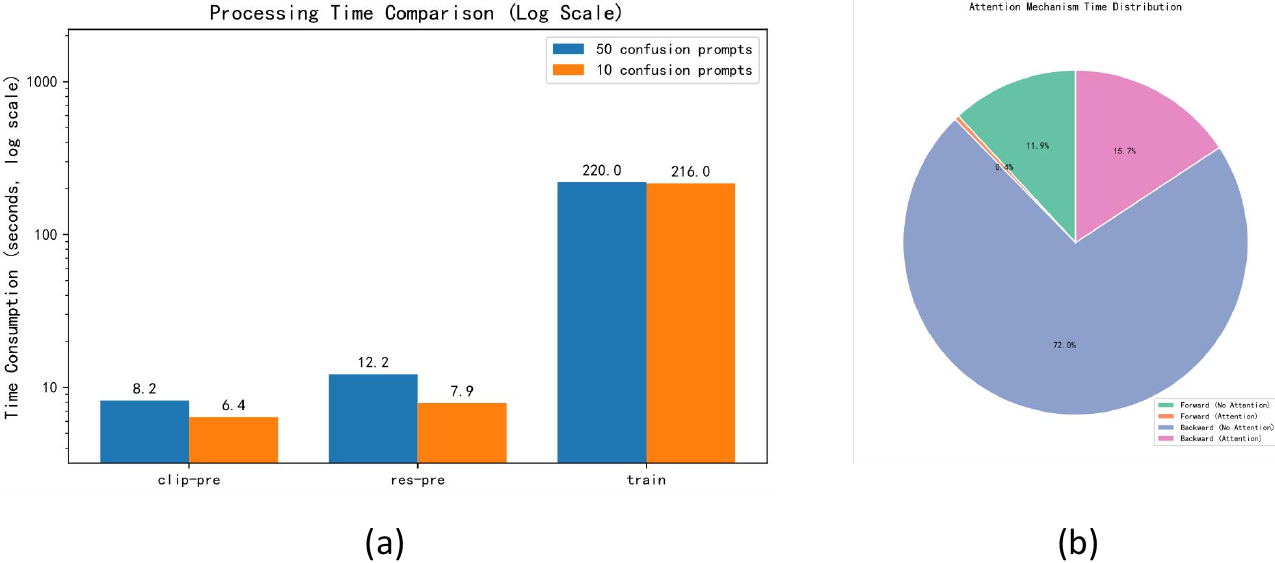}
    \end{minipage}
    
    \caption{(a) Time Comparison Between Introduced Pre-operations and Training Process
(b) Time Consumption of Feature Alignment Module During Training.}
    \label{figure_time}

\end{figure}

\begin{figure}[t]
\begin{minipage}{\linewidth}
    \centering
    \includegraphics[scale=0.4]{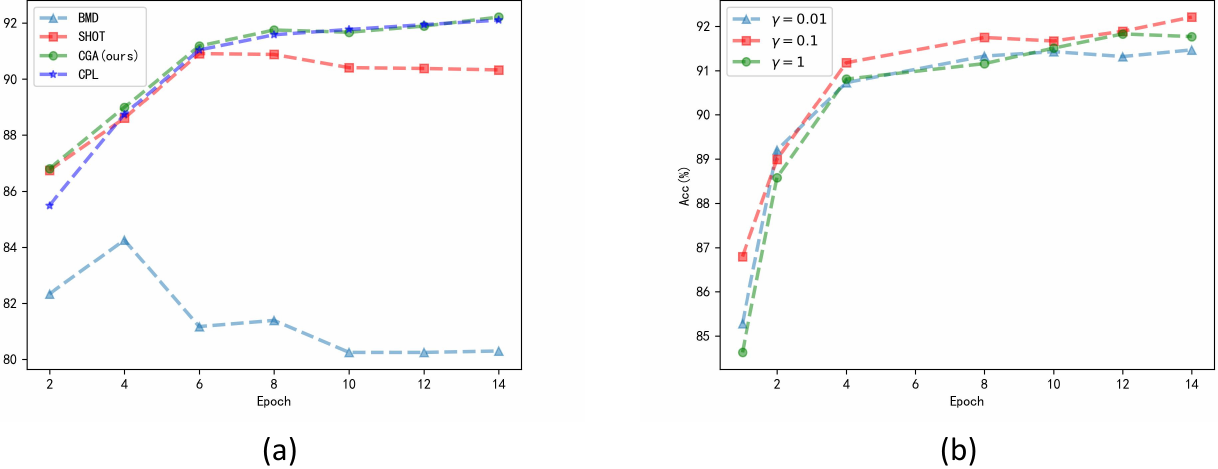}
    \end{minipage}
    
    \caption{(a). The center calculate method; (b) the sensitivity of $\gamma$}
    \label{figure_parameter}

\end{figure}

\begin{figure}[t]
\begin{minipage}{\linewidth}
    \centering
    \includegraphics[scale=0.45]{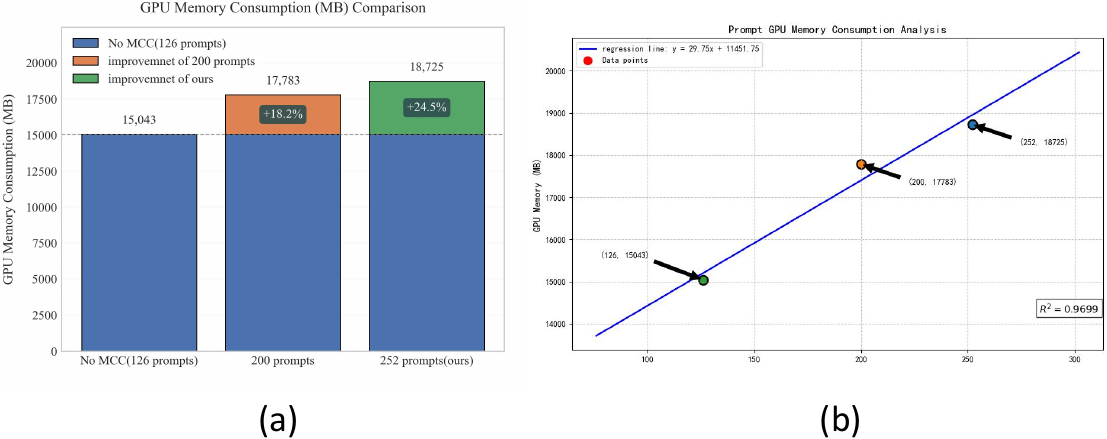}
    \end{minipage}
    
    \caption{Prompt GPU Memory Consumption Analysis.}
    \label{figure_regression}

\end{figure}

In Table \ref{table_ablation-clip}, we compare the improvements brought by our full method and partial methods to CLIP in our experiments.
In the first row of the table, we present the classification accuracy of the original CLIP model across different downstream tasks. Then in the second and third rows, we observed that directly using the Multi-prototype confusion CLIP (MCC) for classification without additional optimization methods actually performed worse. However, simply refining the confusion text  with $L_r$ led to improved classification accuracy over the original CLIP baseline. In the last row , when we applied the complete method, MCC achieved significantly improved results. This demonstrates that our approach not only achieves significant improvements in optimizing the source model, but also effectively enhances the auxiliary CLIP model through mutual learning.

\subsection{Feature distribution visualization.}
In the task of A→C within Office-Home, we use the t-SNE tool~\cite{van2008visualizing} to visualize the feature distribution. And we choose 4 settings, including Source~(only pre-train source model), CLIP~(CLIP'S zero-shot), COWA, and CGA~(ours).

As shown in Fig. \ref{figure_tsne}, from left to right, we observe a progressive reduction in inter-class feature overlap, increasingly compact clustering of intra-class features, achieving the best performance in CGA. This substantiates the efficacy of our method in characterizing feature distributions.
\begin{figure}[H]
\begin{minipage}{\linewidth}
    \centering
    \includegraphics[scale=0.3]{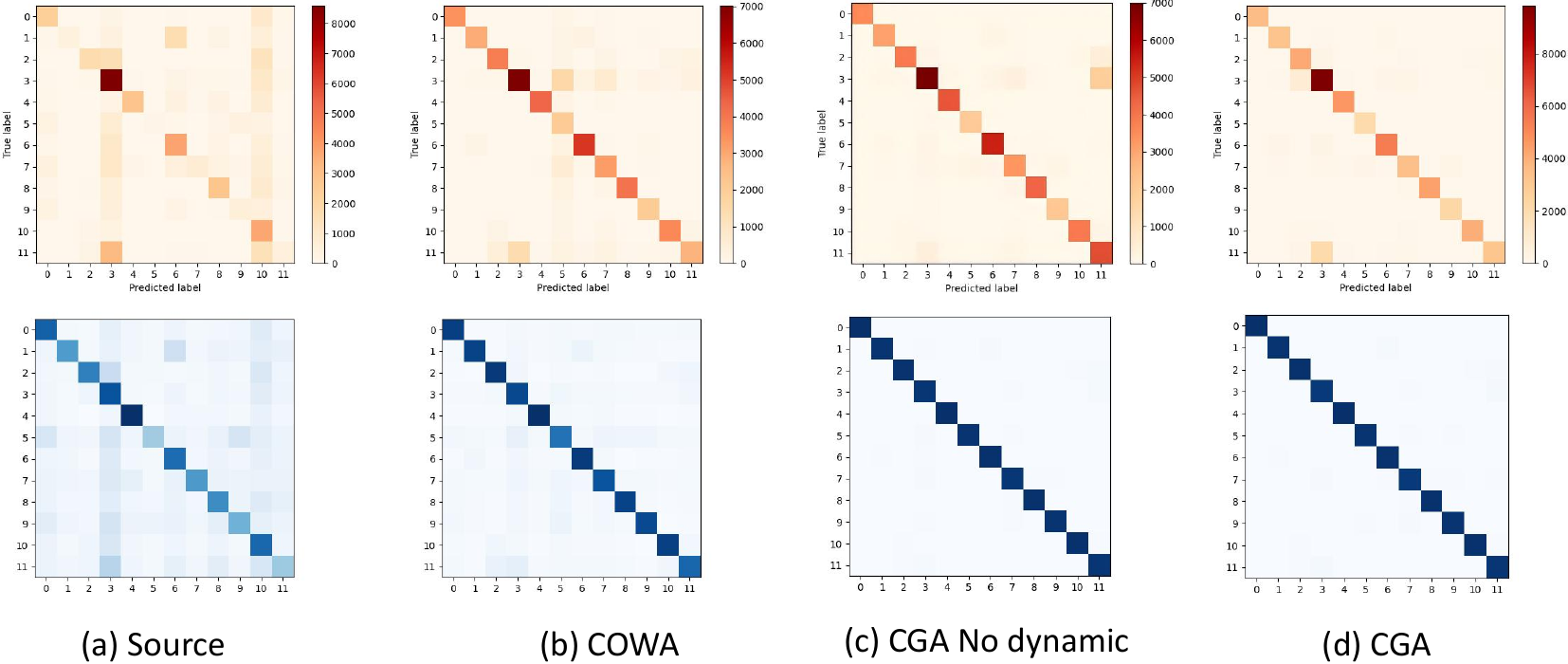}
    \end{minipage}
    
    \caption{The confusion matrix for  classification on VisDA.}
    \label{figure_confusion}

\end{figure}

\subsection{Grad-CAM visualizations.}
We used Grad-CAM~\cite{selvaraju2017grad} visualizations to evaluate the fine-grained discriminative capability of our method trained on Visda across similar categories (bus and truck).
As shown in Fig. \ref{figure_attension}, the source represents the results obtained using a model pre-trained solely on the source domain, while the COWA, CGA w/o MCC and CGA~(ours) show the results from different methods. In the BUS visualization, we can observe that the source method focuses more on the front section of the bus, while COWA pays attention to the overall structure of the bus, and the attention map of CGA w/o MCC indicates that its corresponding model exhibits a broader focus area. In contrast, our CGA method emphasizes the more discriminative doors of the bus. Similarly, in the visualization of truck, both Source and COWA mainly concentrate on the overall structure, particularly the truck cab section,and CGA w/o MCC focus a larger areas, whereas our method targets the distinctive carriage components and undercarriage chassis parts of the truck.

Therefore, our experiments demonstrate that CGA exhibits a strong fine-grained discriminative capacity among similar classes, with a heightened focus on entity-specific distinctive features.

\subsection{Confusion Matrix.}
To quantitatively observe the degree of confusions between classes and to demonstrate the asymmetric classification behavior of our model discussed earlier, we computed the true confusion matrix using ground-truth labels~(the first row) and the estimated confusion matrix~(the second row) via Eq.\ref{2} using pseudo-labels. We conducted experiments on the VisDA using three models: Source, COWA, CGA No dynamic(without dynamically updating confusion relationships and confused class text at each epoch start) and CGA. As shown in Fig. \ref{figure_confusion},
in the first row ,the true confusion matrix, we can observe that while COWA alleviates some of the class confusion present in the pre-trained model to a certain
extent, there are still significant issues between certain classes. In contrast, CGA significantly reduces the class confusion issue, markedly enhancing the fine-grained discriminative capability of the model. Compared with CGA, we can find that CGA No dynamic failed to resolve the confusion between class 3 (car) and class 11 (truck), likely because the initially computed class confusion relationships only captured characteristics from the early training stage. As training progressed and model parameters updated, these fixed relationships became misaligned with the actual model behavior.

Meanwhile, it can be observed that the confusion matrix estimated via Eq.\ref{2} exhibits a notable similarity to the true confusion matrix, particularly in the off-diagonal elements. Furthermore, the weights of the off-diagonal elements in our estimated confusion matrix progressively decrease from left to right, further validating the accuracy of our estimated class confusion relationships and the effectiveness of the proposed method in reducing confusion issues.

Obviously, nearly all confusion matrices, whether true or estimated, are not strict symmetry, which strongly supports our thesis on the model's asymmetric prediction behavior.

\subsection{Parameter Sensitivity.}
Here, we separately discuss the method for calculating the center in Eq.\ref{1} and Eq.\ref{2}, and the influence of the parameter $\gamma$ on the model.
As shown in Fig. \ref{figure_parameter} (a), we compare the setting of effective weight quantity N=2 in our CGA with other methods. We can see the SHOT~(which treats each bit of probability as an effective weight, \textit{i.e.}, N=C), whose performance is suboptimal, likely due to its excessive introduction of interference information, leading to significant errors in both qualitative and quantitative estimation of confused classes. And BMD~(similar to Eq.\ref{10}, selecting only the top M probabilities corresponding to the highest ranked classes with the same weight) perform poorly, because it dismisses the inherently inter-class
confusion patterns in probability. And our method achieves comparable performance to the CPL~\cite{zhang2024candidate}~(which dynamically determines effective weights for each sample) with lower computational complexity.
And in Fig.\ref{figure_parameter}(b), our experiments demonstrate that 0.1 is an appropriate choice for the hyperparameter $\gamma$.

\subsection{Training Time and Memory Usage Analysis.}
In addition to the cross-entropy loss introduced during training, which followed other classification models, we incorporated additional training losses and parameters, resulting in an increased computational overhead in both space and time for our model.
Consequently, we conducted a computational efficiency analysis of the additional components, evaluating both time and space complexity.

First, we conducted a temporal analysis of the pre-processing operations performed by CLIP and the source model at the beginning of each epoch. As illustrated in Fig.\ref{figure_time}(a), experiments under the A-C configuration of the Office-Home dataset demonstrate that the additional time overhead introduced by these two pre-operations remains entirely acceptable relative to the training phase, imposing no significant computational burden. Moreover, the additional time overhead incurred by adding confusion prompts is negligible.

We subsequently analyzed the time overhead introduced by the feature-align module within a single training batch. As evidenced by Fig.\ref{figure_time}(b), the computational latency attributable to the feature alignment module remains marginal, accounting for merely 16\% of the total processing time, during both the forward and backward propagation phases.

Then, we analyze the GPU memory consumption induced by the text prompts generated by Eq.(5). As shown in Fig.\ref{figure_regression} given their participation in training and substantial amount, these prompts incur significant memory overhead. Under the C→P transfer setting on DomainNet-126, the experimental results demonstrate that: The No MCC (using only category prompts) configuration with 126 prompts occupies 15,043 MB of GPU memory. When using 200 prompts, it occupies 17,783 MB, resulting in an additional 18.2\% memory overhead. In contrast, our method (using 252 prompts) occupies 18,725 MB, representing a 24.5\% increase in memory consumption. Additionally, we estimated the average GPU memory consumption per prompt using the regression equation shown in Figure.\ref{figure_regression} b, and the results indicate that each prompt occupies approximately 29 MB of memory on average.

\section{Conclusion}
In this paper,  we propose a targeted guidance approach for CLIP by quantitatively estimating the class confusion relationships of the source model on downstream datasets. Through directional prompt engineering, we generate probabilistic outputs that mitigate existing issues in the source model. Simultaneously, we construct a confused feature center bank to implicitly align the feature spaces between models. This dual mechanism enhances the model's fine-grained discriminative capability among similar classes, further closing the confusion loop in SFDA.

\bibliographystyle{IEEEtran}
\bibliography{IEEE}

\vfill

\end{document}